\documentclass{article}

\PassOptionsToPackage{sort&compress}{natbib} 
\usepackage[preprint]{corl_2026}
\usepackage{booktabs}
\usepackage{multirow}
\usepackage{enumitem}
\usepackage{amsmath}
\usepackage{amssymb}
\usepackage{graphicx}
\usepackage{xcolor}
\usepackage{wrapfig}
\usepackage{pifont}
\usepackage{xspace}
\usepackage{soul}
\usepackage{changes}
\usepackage{cleveref}
\usepackage{pifont}
\usepackage{needspace}
\usepackage[font=normal,labelfont=bf]{caption}

\def\eg{{\it e.g.}\xspace}

\def\ie{{\it i.e.}\xspace}
\def\vs{{\it vs.}\xspace}

\newcommand{\paragraphvA}[1]{\textbf{#1}}

\newcommand{\yes}{\ding{52}}
\newcommand{\no}{\ding{56}}

\newcommand{\TA}{$\mathcal{T}_\text{near}^\text{\yes}$}
\newcommand{\TB}{$\mathcal{T}_\text{far}^\text{\no}$}
\newcommand{\TC}{$\mathcal{T}_\text{near}^\text{\no}$}
\newcommand{\TD}{$\mathcal{T}_\text{far}^\text{\yes}$}
\newcommand{\name}{\textsc{ContactMimic}\xspace}

\PassOptionsToPackage{pagebackref}{hyperref}

\title{\name: \\
Humanoid Object Interaction via Contact Control}

\author{
  Xinyao Li\thanks{Equal contribution.}\qquad
  Xialin He\footnotemark[1]\qquad
  Runpei Dong\qquad
  Saurabh Gupta\\
  University of Illinois Urbana-Champaign
}

\begin{document}
\maketitle

\begin{center}
    \centering
    \captionsetup{type=figure}
    \vspace{-12pt}
    \includegraphics[width=\linewidth]{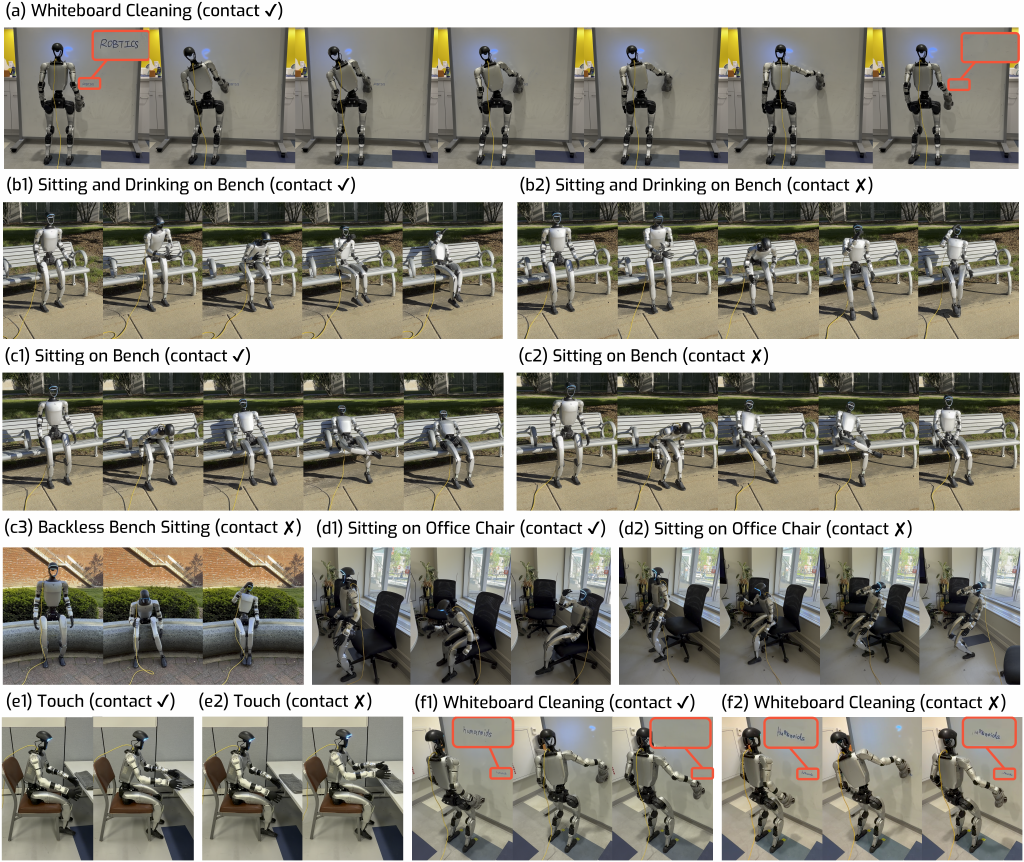}
    \vspace{-8pt}
    \caption{\textbf{\name enables explicit contact control on a real humanoid
    across diverse interaction tasks.} For each task, a same policy is commanded to either make the task-relevant
    contact (\emph{contact}~\ding{52}) or to suppress it
    (\emph{contact}~\ding{56}), by toggling a per-part contact label.
    }
    \label{fig:teaser}
\end{center}

\crefname{section}{\S}{\S}
\crefname{table}{Table}{Tables}
\crefname{figure}{Fig.}{Figs.}

\begin{abstract}
    Keypoint tracking alone is insufficient for object interaction
    tasks such as sitting on a chair, wiping a board, or pushing furniture,
    where the robot can reach the correct pose without making meaningful
    physical contact with the object. We present \name, a learning
    framework that tracks explicit part-level binary contact commands alongside
    keypoint trajectories. \name is made possible through the use of 
    contact-following rewards and a trajectory augmentation scheme aimed at
    breaking the correlations between keypoint trajectories and contact labels.
    The resulting policy successfully decouples contact behavior from keypoint
    geometry, and achieves precise physical contact as well as
    contact-controllability (produce or suppress contact during deployment as
    desired). Simulation experiments across 10 diverse human-object interaction
    motions confirm that \name exhibits contact controllability that enables it
    to complete manipulation tasks without task-specific rewards, while also
    outperforming keypoint-only trackers on contact-relevant tasks. Ablations
    confirm the necessity of the proposed trajectory augmentation scheme and
    sim2real deployment validates contact controllability in the real world
    across 5 different motions. Video results are available on \url{https://lixinyao11.github.io/contactmimic-page/}.
\end{abstract}


\section{Introduction}
\vspace{-8pt}

Most useful whole-body loco-manipulation tasks with a humanoid require making
contacts with the environment in important ways. Consider a humanoid wiping a
whiteboard, or pushing chairs to tidy a room, or picking up boxes. Success is
determined not by the robot's keypoint trajectory, but rather by what body part
contacts what objects, and when. For wiping a whiteboard, it is precisely the
contacts at the hands that differentiate between waving the hand very close to a
whiteboard and actually wiping it. Thus, just keypoint trajectories is an
incomplete specification and many useful loco-manipulation tasks can't be
expressed just with keypoint trajectories. Yet, current humanoid trackers~\cite{beyondmimic,exbody2,omnih2o,humanplus,asap,maskedmimic} aren't
aware of the contacts they should make along the way, and are only trained to
track keypoint trajectories. As a result, while they may reproduce the shape of a
motion, they miss the contacts that make the motion useful for a task. Thus,
even universal keypoint trackers aren't directly useful and require finetuning
with task-specific rewards for success~\cite{beyondmimic,hover,maskedmimic}.

Consider instead \textit{contact-conditioned keypoint trackers}, where rather
than just conditioning trackers on keypoint trajectories, we also provide the
desired per time-step body-link contact labels as additional input to the
policy. The additional contact information disambiguates tasks (\eg wiping \vs
waving, sitting \vs squatting). Furthermore, it can provide more fine-grained
control over the motion, \eg sit in the chair but without leaning on the back.
And finally, it may be possible to train task-specific policies \textit{without
requiring task-specific rewards}, but simply by training for the objective of
matching keypoint and contact trajectories.

So, how do we train a \textit{contact-conditioned keypoint tracker}?  There
are two key considerations: a) how do we source trajectory data with associated
contact labels for training, and b) how to design and train contact conditioned
policies.  For the first question, we leverage human object interaction
datasets. Rather than just extracting and retargeting keypoint trajectories, we
additionally extract and retarget contact patterns. However, in raw human-object interaction data, keypoint trajectories and contact patterns are often strongly correlated: a motion usually appears with only one typical contact pattern. As a result, the policy may ignore the contact command and simply infer the expected contacts from the keypoints.
We mitigate this issue by developing an augmentation scheme that generates
motion pairs that have similar keypoints but differ in the contact
patterns. The policy can't just go by keypoints but has to attend to the
contact commands.

For the second question, we use contact data for policy training in two ways.
First, we inject contact commands into the policy. Second, we include rewards
that encourage the policy to match the commanded contact patterns: a
\emph{contact label matching} reward that measures the agreement between the
reference and actual body-part object-part contacts, and a \emph{contact
distance} reward that guides relevant body parts toward (or away from) the
object surface based on the reference label.

Across both sim and real world experiments with the G1 humanoid robot, we find
that contact controllability works: under the same keypoint trajectory,
changing the control command changes the robot's behavior. This allows us to
execute subtle humanoid object interaction, such as sitting in a chair with or without using the backrest. Compared with a leading
keypoint-only tracker, BeyondMimic~\cite{beyondmimic}, our method achieves more consistent task-relevant contacts and provides explicit contact control. Furthermore,
contact conditioning lets us complete object manipulation tasks (\eg lifting a box),
without needing task specific rewards. Ablations suggest that the specific data
design that breaks the correlations between keypoints and contacts is
important. Finally, we observe that proprioception itself provides cues for
the runtime contact state, justifying our choice of not using contact sensors
as policy inputs.

\section{Related Work}
\label{sec:related}
\vspace{-8pt}

\textbf{Humanoid Whole-Body Tracking and Loco-Manipulation.}
A large body of work trains humanoid policies to imitate human motion through
reinforcement learning, by tracking a purely \emph{kinematic} trajectory of
keypoints or joint
targets~\cite{phc,exbody,exbody2,humanplus,h2o,omnih2o,hover,maskedmimic,asap,beyondmimic,gmt,kungfubot,twist,amo,videomimic}.
Loco-manipulation systems build on such trackers,
driving the upper body via teleoperation, vision, task-specific rewards, or
force adaptation~\cite{homie,idp3,ultra,hero,falcon,visualmimic,opt2skill}.
Across these works, contact with the environment is an incidental byproduct of
matching the reference geometry rather than a controllable target, which breaks
down on tasks whose meaning is defined by contact (\eg wiping a board \vs 
waving just above it). ResMimic~\cite{resmimic} adds a contact-tracking reward
to a residual HOI tracker, but conditions only on motion and object
trajectories and thus doesn't exhibit fine-grained control (\eg sit without
leaning).

\textbf{Human-Object Interaction Data and Retargeting.}
Training contact-aware humanoid policies requires reference data that
also contains contact patterns with objects.
Unlike contact-free motion-capture corpora~\cite{amass}, HOI
datasets~\cite{grab,behave,omomo,humoto} additionally capture object contact.
Retargeting these motions onto a humanoid while preserving contact is
itself non-trivial: naive joint-angle retargeting yields foot-skating,
penetration, or hands floating above intended surfaces.
PHC~\cite{phc} addresses this for free-space motion through
physics-based imitation, while OmniRetarget~\cite{omniretarget} (which we use)
and GMR~\cite{gmr} explicitly preserve interaction structure and reduce
retargeting artifacts. 
Existing pipelines, however, output a single canonical retargeted
trajectory per clip, which is too narrow to teach a policy to
\emph{decouple} contact from keypoint geometry.

\textbf{Contact-Aware Motion Synthesis.}
A separate line of work does model contact explicitly, but not as a
runtime-controllable input to a closed-loop policy.
Classical robotics plans through contact via trajectory optimization with
complementarity constraints~\cite{cio,posa14,tassa12}, and physics-based
character animation conditions motion on scene
contact~\cite{deepmimic,nsm,samp}.
Both require contact schedules to be fixed at planning time.  Learning-based
HOI synthesis methods, whether kinematic
generators~\cite{interdiff,hoidiff,chois,omomo} or physics-based
goal/task-conditioned policies~\cite{sfv,interprior,tokenhsi}, treat contact as
an \emph{intermediate representation} or affordance that keeps motion
physically plausible (\eg hands attached to grasped objects), rather than as an
explicit knob exposed to the user, which is the goal of our work. 

\begin{figure}
\centering
\includegraphics[width=\linewidth]{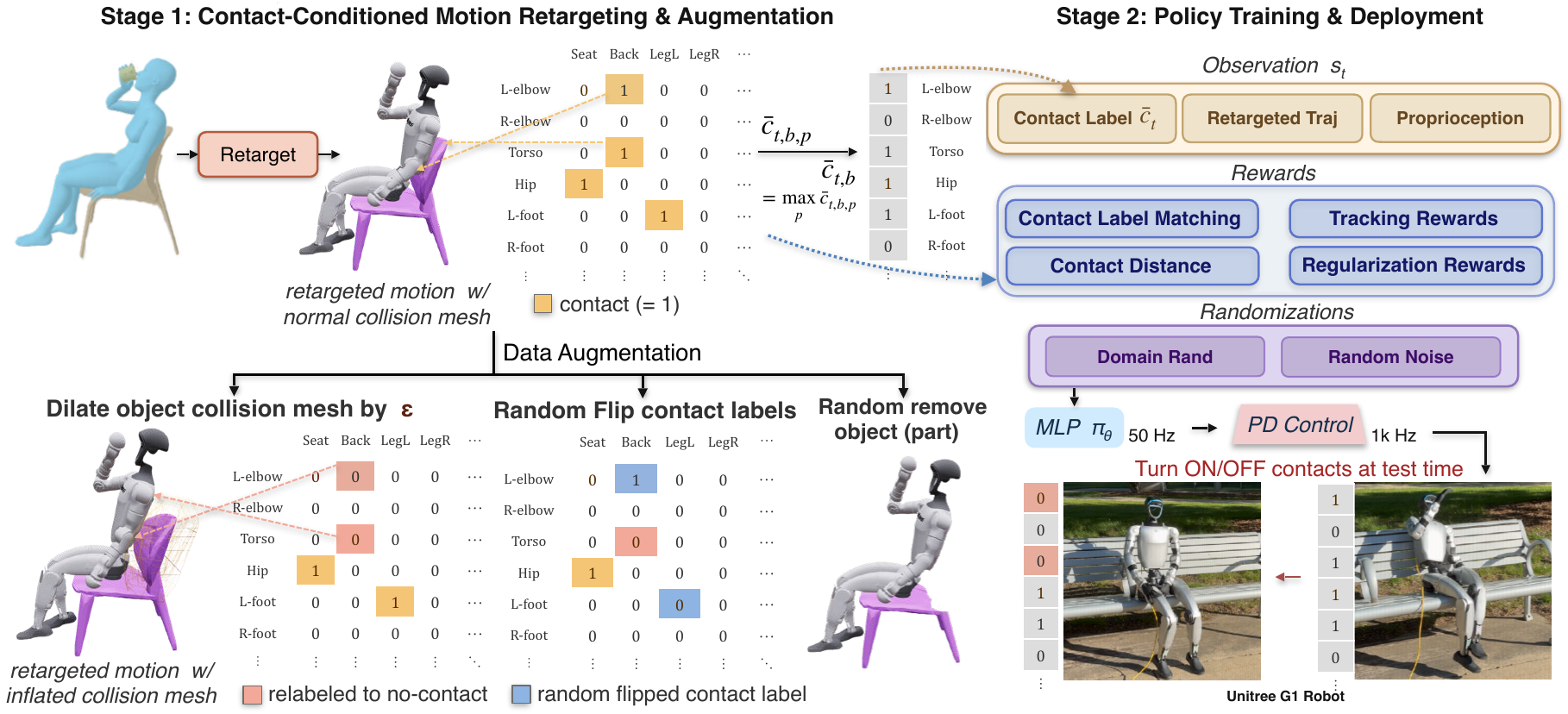}
\vspace{-5pt}
\caption{Overview of our pipeline. We retarget human-object interaction clips
from HUMOTO, extracting both reference keypoint trajectories
and per-body contact labels. We then synthesize augmented motion pairs by \textbf{inflated geometry}, \textbf{contact-label flipping} and \textbf{object removal}. 
A contact-conditioned policy is then trained on this data with contact-and-keypoint-tracking and contact-aware rewards,
so that contact can be \textbf{turned on or off at test time} through the conditioning label.}
\label{fig:pipeline}
\vspace{-10pt}
\end{figure}
\section{Method}
\label{sec:method}
\vspace{-8pt}
Our proposed contact and keypoint tracker, $\pi_\theta(\mathbf{a}_t |
\mathbf{p}_t, \bar{\mathbf{k}}_t, \bar{\mathbf{c}}_t)$, takes as input
proprioception $\mathbf{p}_t$, reference keypoint position targets
$\bar{\mathbf{k}}_t$, and a binary reference contact label map 
$\bar{\mathbf{c}}_t \in \{0,1\}^{|\mathcal{B}|}$, where $\mathcal{B}$ is the
set of contact-capable robot bodies (pelvis, torso, hips, knees, ankles,
shoulders, and wrists).
The policy outputs target joint angles that are converted into torques using a PD controller.

Overall, we follow standard practices in training the policy $\pi$ (see
\cref{fig:pipeline} for an overview): we
pre-process and retarget human-object reference data to generate reference
robot trajectories that are used as tracking targets for training the policy.
Our key technical contributions are in the design of the rewards that encourage
the policy to make the intended contacts (\cref{sec:method_reward}) and in
generating data that allows the policy to learn genuine contact-conditioned
behavior (\cref{sec:data}).
\subsection{Reward Design}
\label{sec:method_reward}
\vspace{-5pt}
The total reward decomposes into standard regularization terms and keypoint-tracking terms inherited from
prior work (described in supp.), and contact-aware rewards introduced in this paper:
\begin{equation}
  r_t = \underbrace{r^\text{track}_t}_{\text{tracking}}
      + \underbrace{w_\text{lm}\, r^\text{lm}_t + w_\text{cd}\, r^\text{cd}_t}_{\text{contact-aware (ours)}}
      + \underbrace{r^\text{reg}_t}_{\text{regularization}}.
\end{equation}
The contact-aware terms compare the reference contact label $\bar{c}_{t,b,p}
\in \{0,1\}$ to the \emph{actual} contact state $c_{t,b,p} \in \{0,1\}$,
between robot body part $b$, object semantic part $p$ at time $t$ via the
following:

\paragraphvA{Contact label assignment,} $r^\text{lm}_t$:
This term rewards the policy for matching $c_{t,b,p}$ to the
reference label $\bar{c}_{t,b,p}$ across all contact pairs.
Let $\mathcal{S}_+ = \{(b,p):\bar{c}_{t,b,p}=1\}$ and
$\mathcal{S}_- = \{(b,p):\bar{c}_{t,b,p}=0\}$ denote the set of object and robot
parts that are / aren't intended to be in contact. Define the True Positive Rate
(TPR), True Negative Rate (TNR), and the False Positive Rate (FPR) as:
\begin{equation}
  \mathrm{\text{TPR}} = \frac{1}{|\mathcal{S}_+|}\!\sum_{(b,p)\in\mathcal{S}_+}\!c_{t,b,p},\quad
  \mathrm{\text{TNR}} = \frac{1}{|\mathcal{S}_-|}\!\sum_{(b,p)\in\mathcal{S}_-}\!(1-c_{t,b,p}),\quad
  \mathrm{\text{FPR}} = 1 - \mathrm{\text{TNR}}.
\end{equation}
$r^\text{lm}_t$ takes either of the following two forms:
\begin{itemize}[leftmargin=1.5em,itemsep=0pt,topsep=2pt]
  \item \textbf{Balanced accuracy} (default):
    $r^\text{lm}_t = \tfrac{1}{2}(\mathrm{\text{TPR}} + \mathrm{\text{TNR}})$.
  \item \textbf{TP$-$FP} (for sparse-contact motions):
    $r^\text{lm}_t = \mathrm{\text{TPR}} - \lambda\,\mathrm{\text{FPR}}$,
    which provides a stronger gradient when most pairs have
    $\bar{c}_{t,b,p}=0$ and TNR saturates.
\end{itemize}

\paragraphvA{Contact distance,} $r^\text{cd}_t$ is computed as 
$r^\text{cd,+}_t + r^\text{cd,-}_t$, where $r^\text{cd,+}_t$ encourages 
part-pairs that are supposed to be in contact to be close, and
$r^\text{cd,-}_t$ encourages part-pairs that are not supposed to be in contact to be
far. Specifically, let $d(b,p)$ denote the distance of the origin of robot body $b$ from the surface of the object
part $p$, and $\mathbf{1}[\cdot]$ the indicator function. $r^\text{cd,+}_t$ and $r^\text{cd,-}_t$ are computed as:
{
\begin{equation}
  r^\text{cd,+}_t = \frac{1}{|\mathcal{S}_+|} \sum_{(b,p)\in \mathcal{S}_+}
    \exp\!\Big( \frac{-d(b, p)^2}{2\sigma^2}\Big), \quad 
    r^\text{cd,-}_t = \frac{-1}{|\mathcal{S}_-|} \sum_{(b,p)\in \mathcal{S}_-}
    \mathbf{1}\!\left[ d(b, p) < \delta \right].
\end{equation}
}



\subsection{Generating Motion Pairs to Break Correlations between Keypoints and Contacts}
\label{sec:data}
\vspace{-5pt}
Our proposed contact and keypoint tracker needs reference trajectories with per time step keypoint and contact labels for training. We extract such
contact label-paired keypoint trajectories from human MoCap data.  
Existing works only extract keypoints, we extend them to also extract contact information. 
As our experiments show, just training on this data alone doesn't lead to good
contact control due to the correlations between keypoint patterns and contacts.
Thus, to achieve genuine contact control, we propose augmentation strategies that break this correlation.

\paragraphvA{Retargeting Human Motions to G1 and Extracting Contact
Information.}
We use OmniRetarget~\cite{omniretarget} to retarget a given human-object interaction clip to the G1 humanoid, producing a reference robot configuration
trajectory $\bar{\mathbf{q}}_{1:T}$ (joint positions and root pose).
We extract \emph{reference contact labels} $\bar{c}_{t,b,p}$ directly
from the retargeted trajectory: at each frame $t$, we mark a robot body part $b$
as in contact ($\bar{c}_{t,b,p} = 1$) when its distance to the object surface is
below a $1$~cm threshold, and assign the semantic object
part $p$ (e.g., chair seat and board surface) as the
nearest object part. 
$\bar{c}_{t,b,p}$ is directly used to compute rewards, and also gives the conditioning vector $\bar{c}_{t,b}$ via $\max_p \bar{c}_{t,b,p}$.

\paragraphvA{Augmented trajectory.}
For each retargeted clip, we synthesize several augmentations that preserve the
overall motion structure but break the correlations between the keypoint
positions and the contact labels for the task relevant contacts (\eg wrist-box
contact for the task of lifting a box). 

Specifically, we generate three different augmentations: 
\ding{182} \textbf{Contact-label flipping}: we retain the original trajectory but flip the task-relevant contact labels, the object remains in simulation but the robot is penalized for making contact with it; 
\ding{183} \textbf{Object removal}: we remove the interaction object and force the target contact labels to zero, while fully retaining the original keypoint trajectory; and 
\ding{184} \textbf{Inflated geometry}: we inflate the collision geometry of target object parts during retargeting to route the robot around them. This produces a perturbed keypoint trajectory where target contacts no longer occur, and those contact labels are zeroed out accordingly. These augmentations are composed together. For instance, combining \textit{Inflated Geometry} with \textit{Contact-label flipping} pairs a distant trajectory with a \textit{true} contact label, simulating the situation where we want contact to happen but reference keypoints are slightly misspecified.

\section{Experiments}
\label{sec:experiments}
\vspace{-8pt}
Our experiments across diverse human-object interaction motions seek
to answer 5 key questions: Do our policies follow the commanded contact
conditioning? Does contact control transfer to the real world? Is keypoint
control alone sufficient, or does a state-of-the-art keypoint-tracking baseline
already produce accurate physical contact? Is our data engineering strategy
necessary for learning contact control? To what extent does the policy's internal
representation already encode the robot's actual runtime contact state, given only
proprioceptive observations?


\subsection{Experimental Setup}
\label{sec:exp_setup}
\setlength{\columnsep}{5pt}

\begin{wraptable}{r}{0.55\textwidth}
\centering
\caption{Motions used in our experiments.
  ``Contact pair'' specifies the task-relevant robot-body and object
  part involved in the contact. All objects, except for the last two motions,
  are fixed in the environment.}
\label{tab:motions}
\setlength{\tabcolsep}{6pt}
\resizebox{\linewidth}{!}{
\begin{tabular}{lllc}
\toprule
\bf Motion & \bf Contact pair & \bf Object (type) \\
\midrule
Wipe whiteboard & hand \& whiteboard surface     & whiteboard  \\
Sit in front of table       & hands \& table top & chair + table  \\
Lean on backrest I       & torso \& chair backrest & chair  \\
Lean on backrest II      & torso \& chair backrest & chair  \\
Step foot on chair              & foot \& chair seat                 & chair  \\
Sit on table            & pelvis / hands \& table top & table  \\
Lean against table     & hands \& table top & table  \\
Sit and squat                  & pelvis \& chair seat (sit only)    & chair  \\
Kick chair            & foot \& chair leg/base             & chair (free) \\
Pick up box         & hands \& box surface               & box (free) \\
\bottomrule
\end{tabular}}
\end{wraptable}

\paragraphvA{Robot and Simulator.}
We use the Unitree G1 humanoid robot (29~DoF) in Isaac Lab~\cite{isaaclab} with PhysX rigid-body simulation.
Each environment contains one interaction object, which is either fixed or free depending on the motion.

\paragraphvA{Motion Dataset.}
We train on 10 motion clips from the HUMOTO~\cite{humoto} dataset, spanning a range of human-object interaction categories, as summarized in \cref{tab:motions}.

\paragraphvA{Metrics.}
Across all motions we report \textbf{contact bodies}, the number of robot bodies
in contact with the \emph{target object part} (e.g., chair seat) averaged over the episode (with the reference count \emph{Ref} for
comparison); \textbf{contact impulse} (N$\cdot$s) accumulated over the episode;
mean \textbf{key-joint torque} (N$\cdot$m) at a motion-specific joint; and
\textbf{MPJPE} (cm) against the reference. For free-object motions we also report
the \textbf{object displacement} (m), how much the object travels from its
initial pose; larger values indicate that the policy
successfully manipulates the object.

\subsection{Contact Control Works in Simulation}
\label{sec:exp_sim}
We measure the extent to which the policy follows the commanded contact. For
the same keypoint trajectory $\tau$, we command the policy to either make
task-relevant contact (denoted by $\tau^\text{\yes}$) or to not make
task-relevant contact (denoted by $\tau^\text{\no}$). For a sitting
motion, $\tau^\text{\yes}$ would be resting the torso against the chair
back, while $\tau^\text{\no}$ would be leaning back but without making
contact with the chair back. We assess whether the policy makes task-relevant
contacts when asked to track $\tau^\text{\yes}$ and whether the
task-relevant contacts drop when asked to track $\tau^\text{\no}$
instead. 
We conduct this test on two sets of keypoint trajectories: a)
$T_\text{kp=near}$, the original object-interaction trajectories in HUMOTO, and
b) $T_\text{kp=far}$, augmented trajectory where the keypoints have been moved
far from the object surface using the inflation process from
\cref{sec:data}. Altogether we test on 4 trajectory sets: \TA, \TB, \TC, \TD.

\begin{wrapfigure}{r}{0.43\textwidth}
  \centering
  \includegraphics[width=\linewidth]{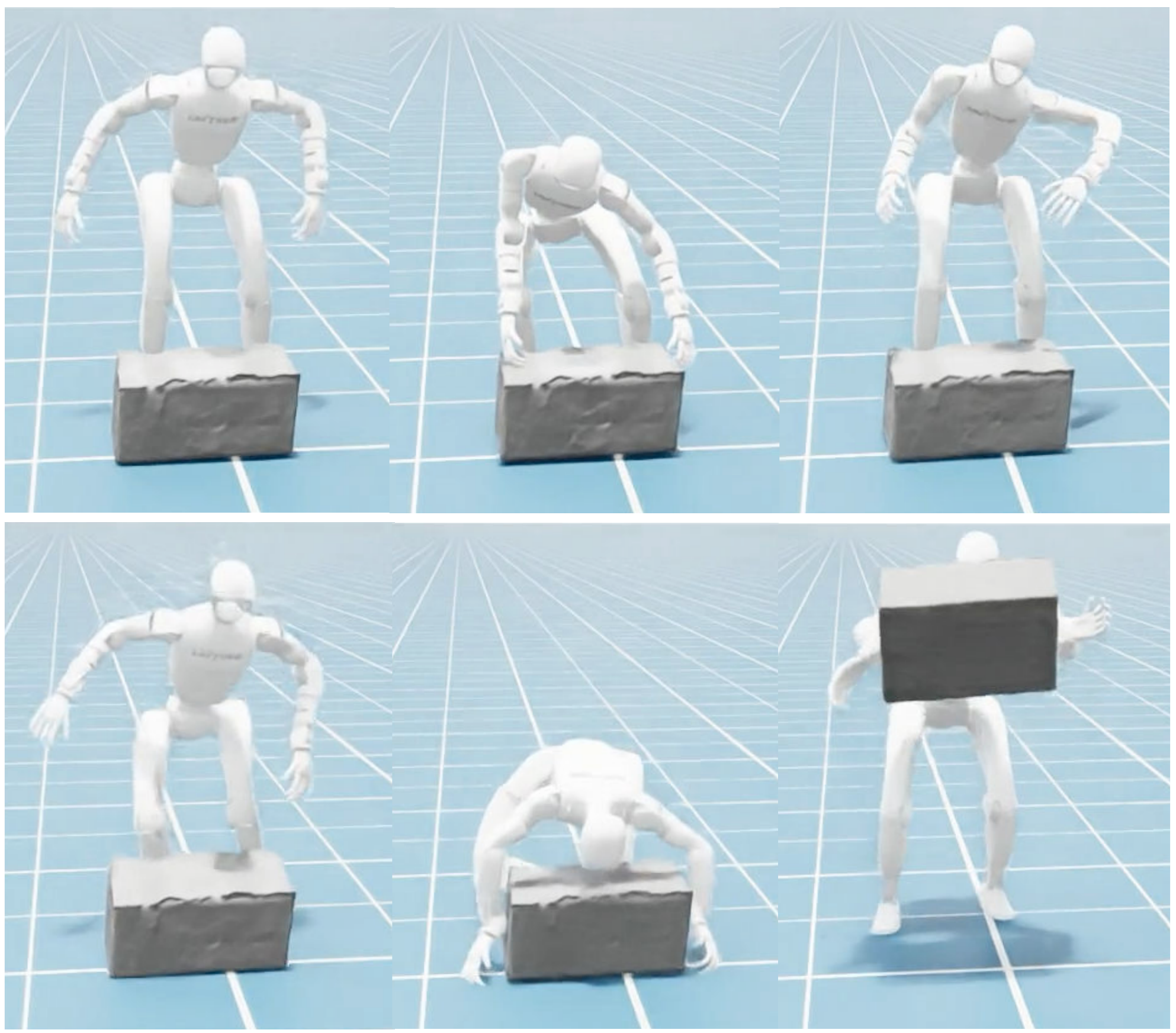}
  \caption{The box-lifting motion in simulation. Top row: contact~\no, the same
    keypoints are tracked but the box is left untouched; bottom row:
    contact~\yes, the robot grasps and lifts the box.}
  \label{fig:box_motion}
\vspace{2mm}
\end{wrapfigure}
\begin{figure}[t]
  \centering
  \includegraphics[width=\linewidth]{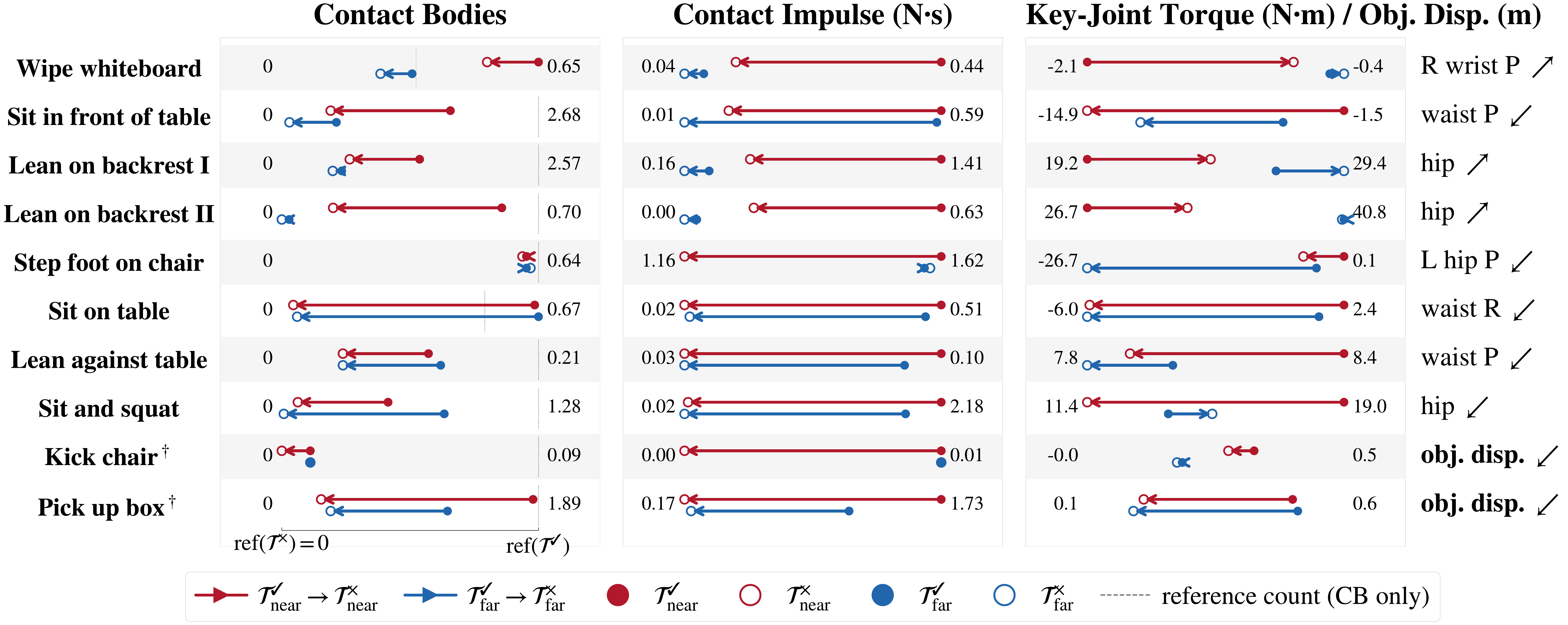}
  \caption{Per-motion visualization of contact controllability.
    \textcolor[HTML]{b2182b}{Red arrows} (near keypoints, contact \yes${\to}$\no, i.e.\ \TA${\to}$\TC)
    and \textcolor[HTML]{2166ac}{blue arrows} (far keypoints, contact \yes${\to}$\no, i.e.\ \TD${\to}$\TB)
    show contact metrics dropping with the contact command turned off.
    Both arrows confirm that the policy corresponds to the
    contact command.
    In the rightmost panel, the arrow beside each key joint marks whether its
    torque should increase ($\nearrow$) or decrease
    ($\swarrow$) when the contact command is turned off.}
  \label{fig:controllability_arrows}
  \vspace{-5mm}
\end{figure}

\textbf{Results.} We plot the number of contacts and contact impulses in
\cref{fig:controllability_arrows}. As expected, we observe that the number of
contacts (and contact impulse) is higher for trajectories \TA and they decrease
as we turn off the contacts in trajectories \TC (the red arrow
points left). The same trend holds when we go from \TD to \TB (the blue arrow
points left). The right pane in \cref{fig:controllability_arrows} shows that the
torque at the contact-relevant changes as expected, \eg the hip torque increases
as the robot supports its back when it is not leaning on the backrest.


For motions that lead to object motion when contact is being made (`Kick chair
with foot' and `Pick up box'), we see that the object displacement is indeed
larger when the policy is asked to make contacts. \cref{fig:box_motion}
visualizes the box lifting motion in simulation. Just with keypoint+contact tracking and without any task-specific rewards, \name is able to loco-manipulate objects.

\vspace{-3mm}
\subsection{Contact Control Transfers to the Real World}
\label{sec:exp_real}
We replicate the contact controllability study on hardware. We evaluate five motions on
the real robot: \emph{wipe whiteboard}, \emph{sit in front of table},
\emph{lean on backrest} (I \& II), and \emph{sit and squat}. For each motion, we deploy the policy with the same keypoints but toggle the contact command from \yes to \no. A trial is successful if the robot qualitatively follows the intended trajectory and its contact behavior matches the  commanded contact, \ie contact between target pair occurs if and only if the contact label is \textit{true} (see \cref{tab:real_results} for per-motion criteria).


\paragraphvA{Results.}
\cref{tab:real_results} reports real-world success rates under both contact commands: \yes and \no. Similar to simulation results, we observe that the policy follows the commanded contact.

\begin{table}[t]
\centering
\caption{Real-world success rates under different controllability conditions. A trial is successful if the robot's physical contact behavior matches contact command (see criteria column).}
\label{tab:real_results}
\setlength{\tabcolsep}{4pt}
\small
\begin{tabular}{llcc}
\toprule
\bf Motion & \bf Success criterion & contact \yes & contact \no \\
\midrule
Wipe whiteboard
  & Hand contact leaves visible trace $\Leftrightarrow$ contact=\yes
  & 5/5 & 5/5 \\
Sit in front of table
  & Full seat contact + hands on table $\Leftrightarrow$ contact=\yes
  & 4/5 & 5/5 \\
Lean on backrest I
  & Sustained torso-backrest contact $\Leftrightarrow$ contact=\yes
  & 9/10 & 10/10 \\
Lean on backrest II
  & Sustained torso-backrest contact $\Leftrightarrow$ contact=\yes
  & 10/10 & 9/10 \\
Sit and squat
  & Seated posture $\Leftrightarrow$ contact=\yes, else squat posture
  & 5/5 & 5/5 \\
\bottomrule
\vspace{-5mm}
\end{tabular}
\end{table}

\paragraphvA{Qualitative observations.}
When commanded to make contact, the policy commits to the physical interaction,
whereas when commanded to suppress it the policy follows a similar pose while
staying just off the surface.
For \emph{wipe whiteboard}, the hand applies sustained pressure, and the eraser erases, \vs hovering just off the board with no effect.
For \emph{sit in front of table}, the robot transfers weight onto the seat and rests its hands on the table top, \vs lowering to the seated pose without bearing weight.
For \emph{lean on backrest}, the torso establishes sustained backrest
contact (with distinct postural styles across the two variants), \vs
remaining upright just shy of the backrest.
For \emph{sit}, the robot transfers weight to the seat, \vs for \emph{squat} it
performs a dynamically balanced squat at the matched height.

\subsection{Keypoint Control Alone Is Insufficient}
\label{sec:exp_necessity}
Next, we assess if just tracking keypoints (the current state-of-the-art) is sufficient when the underlying task requires making contact. Here as a comparison point, we use BeyondMimic~\cite{beyondmimic}, a state-of-the-art whole-body motion tracking framework that doesn't take any contact conditioning as input and is also not trained with any contact rewards. Because BeyondMimic doesn't take any contact conditioning as input, we only train it on sequences that have contact and only compare to it in situations where the task is to make contact.

\paragraphvA{Results.}
\cref{tab:necessity} shows that BeyondMimic~\cite{beyondmimic} has much lower contact metrics (both contact counts and contact impulses) than our method. This suggests that keypoint tracking alone is insufficient for tasks that require making contacts with the environment. At the same time, both methods have comparable MPJPE, indicating that our method is not trading off keypoint tracking accuracy while following contact conditioning. Lastly, for object manipulation tasks, keypoint-only control fails to manipulate the object, while our proposed keypoint+contact control succeeds in manipulating the object, even when the keypoint tracking errors are similar.
\begin{table}[t]
\centering
\caption{Despite similar MPJPE, the keypoint-only method, BeyondMimic~\cite{beyondmimic}, fails to establish physical contact.
  \dag~For object-manipulation motions we also report the object displacement
  (higher means the object is moved more).}
\label{tab:necessity}
\setlength{\tabcolsep}{3pt}
\scriptsize
\newcommand{\std}[1]{{\scriptscriptstyle\pm #1}}
\begin{tabular}{lcccccccc}
\toprule
\multirow{2}{*}{Motion}
  & \multicolumn{2}{c}{Contact bodies$\uparrow$}
  & \multicolumn{2}{c}{Impulse (N$\cdot$s)$\uparrow$}
  & \multicolumn{2}{c}{Obj.\ disp.\ (m)$\uparrow$}
  & \multicolumn{2}{c}{MPJPE (cm)$\downarrow$} \\
\cmidrule(lr){2-3}\cmidrule(lr){4-5}\cmidrule(lr){6-7}\cmidrule(lr){8-9}
  & BM & Ours & BM & Ours & BM & Ours & BM & Ours \\
\midrule
Wipe whiteboard
  & $0.01\std{0.09}$ & $\mathbf{0.65}\std{0.45}$
  & $0.00\std{0.03}$ & $\mathbf{0.44}\std{0.37}$
  & --- & ---
  & $3.9\std{1.0}$ & $3.6\std{0.8}$ \\
Sit in front of table
  & $1.51\std{0.51}$ & $\mathbf{1.76}\std{0.33}$
  & $\mathbf{4.14}\std{8.20}$ & $0.59\std{0.22}$
  & --- & ---
  & $5.8\std{1.9}$ & $5.2\std{2.5}$ \\
Lean on backrest I
  & $0.12\std{0.24}$ & $\mathbf{1.38}\std{1.27}$
  & $0.04\std{0.15}$ & $\mathbf{1.41}\std{1.46}$
  & --- & ---
  & $7.1\std{2.2}$ & $4.8\std{1.3}$ \\
Lean on backrest II
  & $0.39\std{0.35}$ & $\mathbf{0.60}\std{0.39}$
  & $0.59\std{0.58}$ & $\mathbf{0.63}\std{0.55}$
  & --- & ---
  & $4.9\std{1.9}$ & $5.9\std{3.0}$ \\
Step foot on chair
  & $\mathbf{0.62}\std{0.48}$ & $0.61\std{0.48}$
  & $1.29\std{1.06}$ & $\mathbf{1.62}\std{1.33}$
  & --- & ---
  & $3.9\std{1.5}$ & $4.3\std{1.3}$ \\
Sit on table
  & $0.22\std{0.34}$ & $\mathbf{0.66}\std{0.47}$
  & $0.44\std{0.85}$ & $\mathbf{0.51}\std{0.60}$
  & --- & ---
  & $9.5\std{5.0}$ & $9.6\std{3.2}$ \\
Lean against table
  & $0.07\std{0.25}$ & $\mathbf{0.12}\std{0.30}$
  & $0.03\std{0.15}$ & $\mathbf{0.10}\std{0.51}$
  & --- & ---
  & $5.3\std{4.3}$ & $5.9\std{4.0}$ \\
Sit and squat
  & $0.12\std{0.24}$ & $\mathbf{0.53}\std{0.51}$
  & $0.04\std{0.15}$ & $\mathbf{2.18}\std{2.22}$
  & --- & ---
  & $7.1\std{2.2}$ & $5.6\std{1.6}$ \\
Kick chair\dag
  & $\mathbf{0.02}\std{0.09}$ & $0.01\std{0.08}$
  & $\mathbf{0.02}\std{0.23}$ & $0.01\std{0.12}$
  & $0.15\std{0.13}$ & $\mathbf{0.31}\std{0.15}$
  & $4.0\std{1.4}$ & $3.8\std{1.2}$ \\
Pick up box\dag
  & $0.29\std{0.59}$ & $\mathbf{1.85}\std{0.94}$ & $0.09\std{0.22}$ & $\mathbf{1.73}\std{1.18}$ & $0.03\std{0.01}$ & $\mathbf{0.49}\std{0.47}$ & $3.7\std{5.2}$ & $7.5\std{3.5}$ \\
\bottomrule
\vspace{-5mm}
\end{tabular}
\end{table}

\subsection{Our Proposed Data Augmentation Scheme Is Important}
\label{sec:exp_ablation}
\begin{figure}[t]
\centering
\begin{minipage}[c]{0.55\linewidth}
  \centering
  \includegraphics[width=\linewidth]{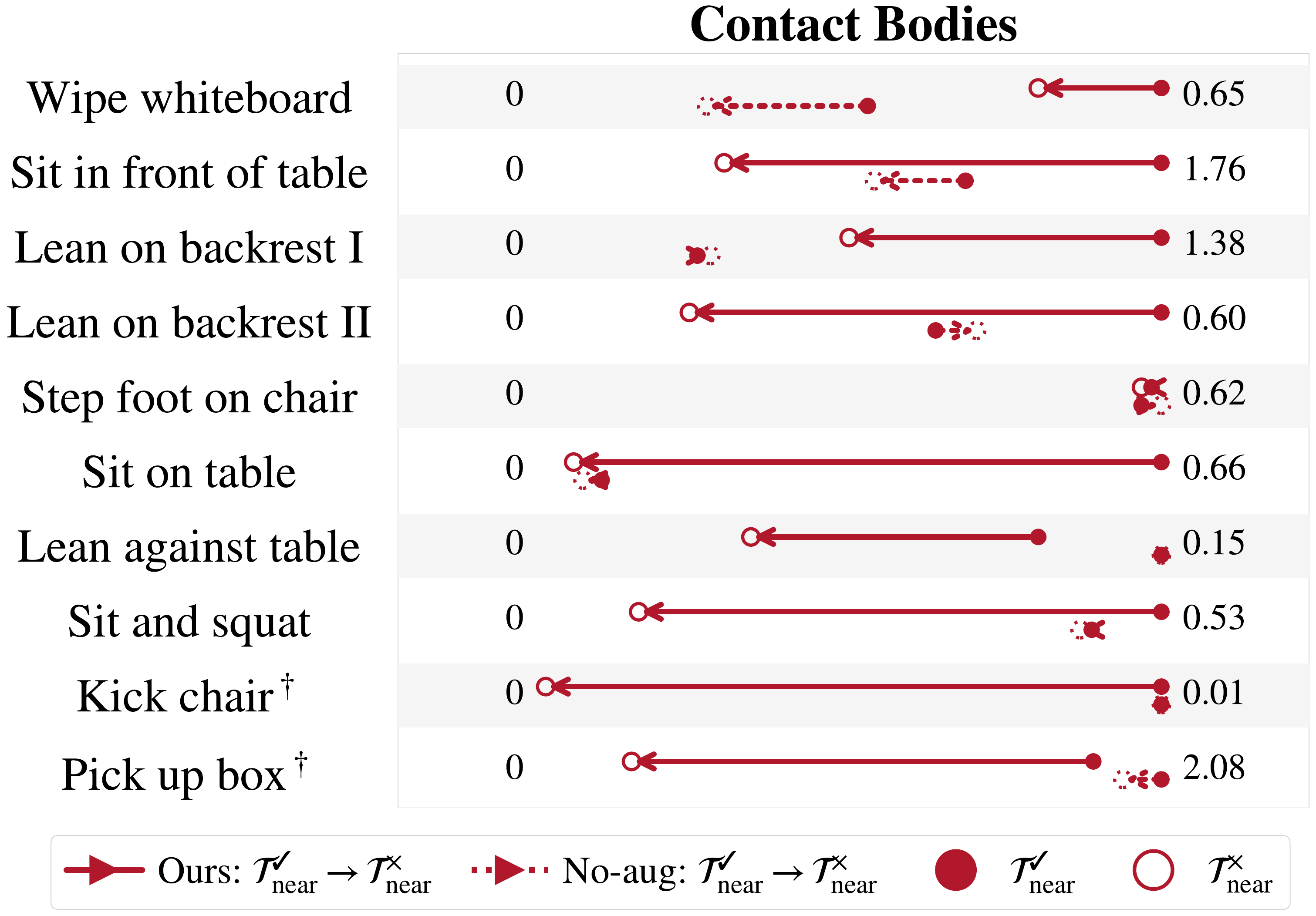}
  \captionof{figure}{\textbf{Removing proposed data augmentation hurts contact controllability.}
    \textcolor[HTML]{b2182b}{Solid arrows} for our full method, are more consistently long and facing left (\ie contacts reduce when commanded not to make contact) than the 
    \textcolor[HTML]{b2182b}{dotted arrows} for the version trained without the augmentations from \cref{sec:data}.}
  \label{fig:ablation_arrows}
\end{minipage}
\hfill
\begin{minipage}[c]{0.43\linewidth}
  \centering
  \setlength{\tabcolsep}{3pt}
  \scriptsize
  \begin{tabular}{lcccc}
  \toprule
  Motion & Cha. & Ref label & Obs & Layer 2 \\
  \midrule
  Wipe whiteboard        & $61$ & $0.843$ & $0.956$ & $\mathbf{0.964}$ \\
  Sit in front of table  & $99$ & $0.762$ & $\mathbf{0.997}$ & $\mathbf{0.997}$ \\
  Lean on backrest I     & $70$ & $0.906$ & $0.943$ & $\mathbf{0.964}$ \\
  Lean on backrest II    & $75$ & $0.894$ & $0.944$ & $\mathbf{0.968}$ \\
  Step foot on chair     & $67$ & $0.963$ & $0.987$ & $\mathbf{0.994}$ \\
  Sit on table           & $69$ & $0.922$ & $\mathbf{0.958}$ & $\mathbf{0.958}$ \\
  Lean against table     & $13$ & $0.885$ & $\mathbf{0.926}$ & $0.915$ \\
  Sit and squat          & $62$ & $0.906$ & $0.970$ & $\mathbf{0.974}$ \\
  Kick chair\dag         & $2$  & $0.000$ & $0.267$ & $\mathbf{0.476}$ \\
  Pick up box\dag        & $81$ & $0.846$ & $0.930$ & $\mathbf{0.931}$ \\
  \bottomrule
  \end{tabular}
  \captionof{table}{\textbf{Runtime contact state can be inferred using
    proprioception.} Linear probing F1 for predicting the runtime contact state
    (task-relevant contact) from the policy input (\emph{Obs}) and layer-2
    representations (\emph{Layer 2}). Both are well above the F1 from the
    reference contact label (\emph{Ref label}); \emph{Cha.}(\emph{Chance} (\%)) is the positive
    fraction of frames.} 
  \label{tab:probe}
\end{minipage}
\vspace{-4mm}
\end{figure}

\vspace{-8pt}
We measure the importance of our data augmentation techniques for successful contact control. \cref{fig:ablation_arrows} compares our method with and without the data augmentation strategies from \cref{sec:data} when tested on trajectories \TA and \TC. As before, left-pointing arrows will suggest good contact control. Our method with data augmentation has much better contact control than the version without the data for all but one motion. This suggests that just the policy input and reward functions aren't sufficient to successfully learn policies with contact-control, training data needs to be properly engineered to break the correlation between keypoint positions and contact labels.




\subsection{Policy Representations Encode Runtime Contact State}
\label{sec:exp_probe}
\vspace{-5pt}
As most robots do not have whole-body contact sensors, we made the deliberate decision to not require the actual runtime contact state for the
different robot body parts as a policy input. Partly our rationale was that the
contact state could be inferred using proprioception. We assess this
quantitatively by computing the contact state prediction accuracy of a linear
probe on the policy inputs and the policy's intermediate representations.


\paragraphvA{Results.}
\cref{tab:probe} reports the test-set F1 scores across all motions.
 For both the raw observation and Layer 2 representations, the F1 scores are very high, and much higher than the chance performance (\ie the number of time steps with contact) as well as the reference label itself. 
Thus, despite receiving no explicit contact sensing at inference time, the policy has a sense of the robot's runtime contact state and
can adapt its behavior to fulfill the commanded contact state.



\section{Discussion and Limitations}
\label{sec:limitations}
\vspace{-8pt}
We presented \name, a framework that augments humanoid keypoint tracking with
an explicit, test-time controllable contact label. By training a single policy
on paired motions that share keypoint structure but differ in contact commands,
and supervising it with contact-aware rewards, the policy learns to make or
suppress physical contact on command.  Across diverse human-object interaction
motions in simulation and on the real Unitree G1, the policy follows the
commanded contact, outperforms a keypoint-only baseline on contact metrics at
comparable tracking accuracy, and even encodes its runtime contact state from
proprioception alone. We believe explicit contact conditioning is a simple and
general interface for expressing contact-rich tasks, and a step toward
humanoids that interact with the world purposefully rather than incidentally.

Our approach has several limitations. First, we train a separate policy per
motion rather than a single universal contact-conditioned tracker. Jointly
training a contact-conditioned policy across motions would be a natural next
step. Second, we built our approach on high-quality human-object interaction
data from the HUMOTO dataset which limits the diversity of interactions
we can currently cover. Developing methods that can use in-the-wild video data
instead can help. Finally, our real-world evaluation spans five motions on a
single robot; broader hardware validation is future work.


%



\clearpage
\acknowledgments{We are grateful to the Coordinated Science Laboratory and the Center for Autonomy for access to experimental space and the G1 humanoid robot used in this work. We thank John Hart for his invaluable help with the humanoid robots. This material is based upon work supported by an NSF CAREER Award (IIS-2143873).}


\bibliography{main}  

\clearpage
\appendix
\newcommand{\std}[1]{{\scriptscriptstyle\pm #1}}

\begin{center}
{\Large\scshape Appendix}
\end{center}

\noindent We provide additional details and results
that supplement \name:
\begin{enumerate}[leftmargin=2em,itemsep=2pt,topsep=2pt]
  \item \cref{sec:appendix_videos} describes the organization of the overview video and the project page.
  \item \cref{sec:implementation} provides additional implementation details,
    including PPO hyperparameters, policy observations and architecture, and
    domain randomization.
  \item \cref{sec:appendix_rewards} lists the full set of reward terms with
    their weights and shaping parameters.
  \item \cref{sec:appendix_data} details our paired-motion generation
    pipeline: contact-label extraction, augmentation parameters, and the
    per-motion training configuration.
  \item \cref{sec:appendix_real} describes our real-world deployment on the
    Unitree G1.
  \item \cref{sec:appendix_results} reports additional quantitative results,
    including the full per-motion contact controllability table and the
    per-motion data augmentation ablation.
\end{enumerate}

\section{Overview Video and Project Page}
\label{sec:appendix_videos}
We provide an overview video and the raw recordings of every real-world trial reported in \cref{tab:real_results} on our \href{https://lixinyao11.github.io/contactmimic-page/}{project page}.
The videos and page include:
\begin{itemize}[leftmargin=1.5em,itemsep=0pt,topsep=2pt]
  \item \textbf{Method overview}: the full pipeline, from human-motion
    retargeting and paired-motion augmentation that breaks keypoint--contact
    correlations, to contact-conditioned policy training with our
    contact-following rewards and real-world deployment.
  \item \textbf{Real-world results}: for each of the five tested motions
    (\emph{wipe whiteboard}, \emph{sit in front of table}, \emph{lean on
    backrest} I/II, \emph{sit and squat}), side-by-side clips of the same policy
    on the real Unitree G1 commanded with contact~\yes and contact~\no,
    demonstrating on-command toggling of physical contact.
  \item \textbf{Simulation results}: per-motion clips across all 10 motions
    under the four eval-time trajectories \TA--\TD, including the flipped
    conditions \TC and \TD that test pure contact-label following.
  \item \textbf{Baseline comparison}: side-by-side comparisons against the
    state-of-the-art keypoint-only tracker BeyondMimic~\cite{beyondmimic}
    under the same keypoint trajectory, showing that without contact
    conditioning it can only follow the single mode baked into the trajectory
    and cannot be steered at test time.
  \item \textbf{Ablation}: comparisons against a no-augmentation variant,
    showing that without our paired-motion augmentation the policy ignores the
    contact command and fails contact-controlled tasks (e.g., \emph{sit on
    table}, \emph{lean on backrest II}).
\end{itemize}

\section{Implementation Details}
\label{sec:implementation}
We train one policy per motion with PPO~\cite{ppo} in Isaac
Lab~\cite{isaaclab} across $4096$ parallel environments at $50$\,Hz. At each environment reset, the
episode is randomly assigned either the default trajectory or one of its
augmented counterparts (\cref{sec:data}), so the same policy sees both contact
and no-contact target labels during training. PPO hyperparameters are listed in
\cref{tab:hyperparams}.

\begin{table}[h]
\centering
\caption{PPO hyperparameters used to train each contact-conditioned policy.}
\label{tab:hyperparams}
\small
\setlength{\tabcolsep}{8pt}
\begin{tabular}{ll}
\toprule
\bf Parameter & \bf Value \\
\midrule
Parallel environments & $4096$ \\
Policy step rate & $50$\,Hz \\
Steps per env per update & $24$ \\
Max iterations & $30{,}000$ \\
Learning rate & $1\!\times\!10^{-3}$ (adaptive, KL target $0.01$) \\
Discount factor $\gamma$ & $0.99$ \\
GAE $\lambda$ & $0.95$ \\
PPO clip parameter & $0.2$ \\
Mini-batches per epoch & $4$ \\
Learning epochs per update & $5$ \\
Entropy coefficient & $5\!\times\!10^{-3}$ \\
Value-loss coefficient & $1.0$ \\
Max gradient norm & $1.0$ \\
Initial action noise std & $1.0$ \\
\bottomrule
\end{tabular}
\end{table}

\paragraphvA{Policy Observations.}
The actor observation $\mathbf{o}_t$ concatenates:
(i) \emph{proprioception} $\mathbf{p}_t$: joint positions, joint velocities,
base angular velocity, and projected gravity;
(ii) \emph{reference keypoint targets} $\bar{\mathbf{k}}_t$, expressed in the
robot's local frame and derived from the reference configuration
$\bar{\mathbf{q}}_t$ via forward kinematics;
(iii) the \emph{binary reference contact label}
$\bar{\mathbf{c}}_t \in \{0,1\}^{|\mathcal{B}|}$
(with $\bar{c}_{t,b} = \max_p \bar{c}_{t,b,p}$) for the current frame.
Uniform observation noise is added during training to all actor observations.

\paragraphvA{Policy Architecture.}
Both actor and critic are MLPs with hidden dimensions $[512, 256, 128]$ and ELU
activations. The critic receives \emph{noise-free} versions of the actor's
observations, plus the base linear velocity, which is not available to the
actor; this acts as the critic's privileged information.

\paragraphvA{Domain Randomization.}
We randomize, at each episode start, robot link masses, joint friction, and
object friction and mass within standard ranges, to facilitate sim-to-real
transfer.

\section{Reward Details}
\label{sec:appendix_rewards}
Following \cref{sec:method_reward}, the per-step reward decomposes into
tracking, contact-aware, and regularization terms. \cref{tab:rewards} lists
every reward term, its functional form, weight, and any tunable shaping
parameter. Tracking terms follow BeyondMimic~\cite{beyondmimic} and use Gaussian
kernels $\exp(-e^2/\sigma^2)$ on the corresponding error signal $e$.

\begin{table}[h]
\centering
\caption{Full list of reward terms used to train each contact-conditioned
  policy. Positive weights are rewards; negative weights are penalties.
  Gaussian terms are shaped as $\exp(-e^2/\sigma^2)$ on the underlying error
  $e$.}
\label{tab:rewards}
\footnotesize
\setlength{\tabcolsep}{5pt}
\begin{tabular}{llp{0.42\linewidth}c}
\toprule
\bf Group & \bf Term & \bf Error signal / form & \bf Weight \\
\midrule
\multirow{6}{*}{Tracking}
  & Global anchor pos. & Root position error; $\sigma\!=\!0.3$\,m       & $0.5$ \\
  & Global anchor ori. & Root orientation error; $\sigma\!=\!0.4$\,rad  & $0.5$ \\
  & Body position      & Per-body position error; $\sigma\!=\!0.3$\,m   & $1.0$ \\
  & Body orientation   & Per-body orientation error; $\sigma\!=\!0.4$\,rad & $1.0$ \\
  & Body lin.\ vel.    & Per-body linear velocity error; $\sigma\!=\!1.0$\,m/s & $1.0$ \\
  & Body ang.\ vel.    & Per-body angular velocity error; $\sigma\!=\!\pi$\,rad/s & $1.0$ \\
\midrule
\multirow{2}{*}{Contact}
  & Label matching     & Balanced accuracy $\tfrac{1}{2}(\mathrm{TPR}{+}\mathrm{TNR})$, force threshold $1$\,N & $4.0$ \\
  & Distance           & Gaussian on body-to-part dist., $\sigma\!=\!0.2$\,m; unwanted-contact penalty at $\delta\!=\!0.05$\,m & $3.0$ \\
\midrule
\multirow{3}{*}{Reg.}
  & Action rate        & $\|\mathbf{a}_t - \mathbf{a}_{t-1}\|_2^2$            & $-0.15$ \\
  & Joint limits       & Quadratic, on joint-limit violation                  & $-10.0$ \\
  & Undesired contacts & Non-foot/wrist body in contact with terrain          & $-0.1$ \\
\bottomrule
\end{tabular}
\end{table}

The contact label-matching reward defaults to the balanced-accuracy form
$r^\text{lm}_t = \tfrac{1}{2}(\mathrm{TPR}+\mathrm{TNR})$; for sparse-contact
motions we instead use the TP$-$FP form $r^\text{lm}_t = \mathrm{TPR} - \lambda\,\mathrm{FPR}$
with $\lambda = 1.0$, which provides a stronger gradient when most
$\bar{c}_{t,b,p}=0$ and TNR saturates.

\section{Paired Motion Generation Details}
\label{sec:appendix_data}
\paragraphvA{Contact label extraction.}
Given a retargeted trajectory from OmniRetarget~\cite{omniretarget}, we replay
the trajectory in MuJoCo and at each frame $t$ mark a robot body part $b$ as in
contact ($\bar{c}_{t,b,p}=1$) when its distance to the object surface is below
$1$\,cm; the semantic part $p$ is assigned by looking up the nearest object
surface point.

\paragraphvA{Augmentation parameters.}
We synthesize three augmentations (\cref{sec:data}). For inflated-geometry
augmentation, the collision geometry of the target object part(s) is expanded
outward by an isotropic offset of $\delta_\text{infl} \in [5, 10]$\,cm in all directions during a second retargeting pass; this forces the retargeter to
route the relevant robot bodies further from the original surface, and the
contact labels of the target pairs are zeroed out accordingly. At training time we further apply
random contact-label flipping and random interaction-object removal; the
per-motion probabilities vary slightly and are listed in
\cref{tab:per_motion_settings}.

\paragraphvA{Per-motion settings.}
\cref{tab:per_motion_settings} summarizes the per-motion training
configuration: which augmentations are used, the contact label-matching
reward mode (balanced accuracy or TP$-$FP), the augmentation probabilities,
and the number of PPO iterations. Sparse-contact motions (e.g., \emph{wipe
whiteboard}, free-object kicks and lifts) use the TP$-$FP mode to provide a
stronger gradient when most reference labels are zero.

\begin{table}[h]
\centering
\caption{Per-motion training settings. \emph{Aug.} columns indicate which
  paired-motion augmentations are enabled: \emph{Infl.}\ (inflated geometry),
  \emph{Rem.}\ (interaction-object removal at train time), and \emph{Flip}
  (random contact-label flipping). \emph{Mode} is the label-matching reward
  form (Bal.\ = balanced accuracy, TP$-$FP = $\mathrm{TPR}-\lambda\mathrm{FPR}$,
  $\lambda{=}1$). $p_\text{rem.}$ and $p_\text{flip}$ are the train-time
  augmentation probabilities (per episode).}
\label{tab:per_motion_settings}
\footnotesize
\setlength{\tabcolsep}{4pt}
\begin{tabular}{lccccc}
\toprule
\bf Motion & \bf Infl.\ aug. & \bf Rem.\ aug. & \bf Flip aug. & \bf Mode & $p_\text{rem.}/p_\text{flip}$ \\
\midrule
Wipe whiteboard         & \yes & \yes & \yes & TP$-$FP & $0.3$ / $0.2$ \\
Sit in front of table   & \yes & \yes & \yes & Bal.    & $0.3$ / $0.2$ \\
Lean on backrest I      & \yes & \yes & \yes & Bal.    & $0.3$ / $0.2$ \\
Lean on backrest II     & \yes & \yes & \yes & Bal.    & $0.3$ / $0.2$ \\
Step foot on chair      & \yes & \yes & \yes & Bal.    & $0.3$ / $0.2$ \\
Sit on table            & \yes & \no & \yes & Bal.    & $0.0$ / $0.2$ \\
Lean against table      & \yes & \yes & \yes & Bal.    & $0.3$ / $0.2$ \\
Sit and squat           & \no  & \yes & \yes & Bal.    & $0.3$ / $0.2$ \\
Kick chair\dag          & \yes & \yes & \yes & TP$-$FP & $0.3$ / $0.2$ \\
Pick up box\dag         & \yes & \yes & \yes & TP$-$FP & $0.3$ / $0.2$ \\
\bottomrule
\end{tabular}
\end{table}

\emph{Sit and squat} is the only motion that does \emph{not} use inflated
geometry: because the augmented ``no-contact'' trajectory for this motion is
simply a squat at the same height, we obtain it directly by removing the chair
from the scene rather than by re-retargeting against an inflated chair.
\emph{Sit on table} skips
the object-removal augmentation since the robot needs to sit on the table surface, and removing the table would directly lead to failure.

\section{Real-World Deployment}
\label{sec:appendix_real}
We deploy each policy on a Unitree G1 ($29$ actuated DoFs). The policy runs
onboard at $50$\,Hz, producing target joint positions that are tracked by the
robot's joint-level PD controller at $1000$\,Hz, identical to the simulation
control stack. Object placement (chair, table, board, bench) is measured by
hand before each trial and reproduced on a fixed mat to align the robot's start
pose with the reference. No external motion capture or vision is used at
deployment time; the policy receives only its onboard proprioception, the
pre-recorded reference, and the contact-label command.

\section{Additional Quantitative Results}
\label{sec:appendix_results}

\paragraphvA{Per-motion contact controllability.}
Table~\ref{tab:controllability} reports the full per-motion contact controllability results across all four trajectories (\TA--\TD). Contact bodies and impulse track the contact label rather than the keypoint variant, confirming the policy follows the commanded contact.

\begin{table}[h]
\centering
\caption{Contact label controllability across all interaction motions.
  A single contact-conditioned policy is evaluated on four input trajectories
  per motion: \TA and \TB are matched (keypoint and label agree),
  while \TC and \TD are flipped (keypoint and label disagree).
  Object displacement is only reported for free-object motions (\dag).
  Bodies/impulse track the contact label (not the keypoint variant),
  indicating the policy follows the label rather than implicitly inferring
  contact from keypoints.
  The arrow beside each key joint marks whether its torque is intuitively
  expected to increase ($\uparrow$) or decrease ($\downarrow$) when the
  commanded contact is turned off.}
\label{tab:controllability}
\setlength{\tabcolsep}{3pt}
\footnotesize
\begin{tabular}{llcccc}
\toprule
Motion & Trajectory
  & Contact bodies
  & Ref
  & Impulse (N$\cdot$s)
  & Key torque (N$\cdot$m) \\
\midrule
\multirow{4}{*}{Wipe whiteboard}
  & \TA & $0.65{\scriptstyle\pm 0.45}$ & $0.34$ & $0.44{\scriptstyle\pm 0.37}$ & $-2.07{\scriptstyle\pm 1.61}$ (R wrist pitch $\uparrow$) \\
  & \TB & $0.25{\scriptstyle\pm 0.25}$ & $0.00$ & $0.04{\scriptstyle\pm 0.06}$ & $-0.44{\scriptstyle\pm 0.30}$ (R wrist pitch $\uparrow$) \\
  & \TC & $0.52{\scriptstyle\pm 0.44}$ & $0.34$ & $0.12{\scriptstyle\pm 0.13}$ & $-0.76{\scriptstyle\pm 0.60}$ (R wrist pitch $\uparrow$) \\
  & \TD & $0.33{\scriptstyle\pm 0.39}$ & $0.00$ & $0.07{\scriptstyle\pm 0.11}$ & $-0.53{\scriptstyle\pm 0.46}$ (R wrist pitch $\uparrow$) \\
\addlinespace
\multirow{4}{*}{Sit in front of table}
  & \TA & $1.76{\scriptstyle\pm 0.33}$ & $2.68$ & $0.59{\scriptstyle\pm 0.22}$ & $-1.46{\scriptstyle\pm 6.26}$ (waist pitch $\downarrow$) \\
  & \TB & $0.08{\scriptstyle\pm 0.19}$ & $0.00$ & $0.01{\scriptstyle\pm 0.09}$ & $-12.12{\scriptstyle\pm 5.08}$ (waist pitch $\downarrow$) \\
  & \TC & $0.51{\scriptstyle\pm 0.64}$ & $2.68$ & $0.11{\scriptstyle\pm 0.21}$ & $-14.91{\scriptstyle\pm 7.54}$ (waist pitch $\downarrow$) \\
  & \TD & $0.57{\scriptstyle\pm 0.46}$ & $0.00$ & $0.58{\scriptstyle\pm 2.70}$ & $-4.65{\scriptstyle\pm 8.78}$ (waist pitch $\downarrow$) \\
\addlinespace
\multirow{4}{*}{Lean on backrest I}
  & \TA & $1.38{\scriptstyle\pm 1.27}$ & $2.57$ & $1.41{\scriptstyle\pm 1.46}$ & $19.2{\scriptstyle\pm 8.4}$ (hip $\uparrow$) \\
  & \TB & $0.51{\scriptstyle\pm 0.49}$ & $1.36$ & $0.16{\scriptstyle\pm 0.27}$ & $29.4{\scriptstyle\pm 17.3}$ (hip $\uparrow$) \\
  & \TC & $0.68{\scriptstyle\pm 0.65}$ & $2.57$ & $0.48{\scriptstyle\pm 0.64}$ & $24.1{\scriptstyle\pm 12.5}$ (hip $\uparrow$) \\
  & \TD & $0.60{\scriptstyle\pm 0.48}$ & $1.36$ & $0.28{\scriptstyle\pm 0.26}$ & $26.7{\scriptstyle\pm 13.3}$ (hip $\uparrow$) \\
\addlinespace
\multirow{4}{*}{Lean on backrest II}
  & \TA & $0.60{\scriptstyle\pm 0.39}$ & $0.70$ & $0.63{\scriptstyle\pm 0.55}$ & $26.7{\scriptstyle\pm 12.0}$ (hip $\uparrow$) \\
  & \TB & $0.00{\scriptstyle\pm 0.02}$ & $0.00$ & $0.00{\scriptstyle\pm 0.06}$ & $40.7{\scriptstyle\pm 18.9}$ (hip $\uparrow$) \\
  & \TC & $0.14{\scriptstyle\pm 0.24}$ & $0.70$ & $0.17{\scriptstyle\pm 0.37}$ & $32.2{\scriptstyle\pm 16.6}$ (hip $\uparrow$) \\
  & \TD & $0.02{\scriptstyle\pm 0.11}$ & $0.00$ & $0.03{\scriptstyle\pm 0.27}$ & $40.8{\scriptstyle\pm 21.1}$ (hip $\uparrow$) \\
\addlinespace
\multirow{4}{*}{Step foot on chair}
  & \TA & $0.61{\scriptstyle\pm 0.48}$ & $0.64$ & $1.62{\scriptstyle\pm 1.33}$ & $+0.10{\scriptstyle\pm 11.62}$ (L hip pitch $\downarrow$) \\
  & \TB & $0.62{\scriptstyle\pm 0.48}$ & $0.00$ & $1.60{\scriptstyle\pm 1.31}$ & $-26.66{\scriptstyle\pm 19.59}$ (L hip pitch $\downarrow$) \\
  & \TC & $0.60{\scriptstyle\pm 0.48}$ & $0.64$ & $1.16{\scriptstyle\pm 1.01}$ & $-4.13{\scriptstyle\pm 14.82}$ (L hip pitch $\downarrow$) \\
  & \TD & $0.61{\scriptstyle\pm 0.48}$ & $0.00$ & $1.59{\scriptstyle\pm 1.32}$ & $-2.77{\scriptstyle\pm 11.17}$ (L hip pitch $\downarrow$) \\
\addlinespace
\multirow{4}{*}{Sit on table}
  & \TA & $0.66{\scriptstyle\pm 0.47}$ & $0.53$ & $0.51{\scriptstyle\pm 0.60}$ & $+2.43{\scriptstyle\pm 5.29}$ (waist roll $\downarrow$) \\
  & \TB & $0.04{\scriptstyle\pm 0.18}$ & $0.00$ & $0.03{\scriptstyle\pm 0.25}$ & $-5.99{\scriptstyle\pm 4.97}$ (waist roll $\downarrow$) \\
  & \TC & $0.03{\scriptstyle\pm 0.16}$ & $0.53$ & $0.02{\scriptstyle\pm 0.20}$ & $-5.91{\scriptstyle\pm 5.01}$ (waist roll $\downarrow$) \\
  & \TD & $0.67{\scriptstyle\pm 0.46}$ & $0.00$ & $0.48{\scriptstyle\pm 0.43}$ & $+1.61{\scriptstyle\pm 5.29}$ (waist roll $\downarrow$) \\
\addlinespace
\multirow{4}{*}{Lean against table}
  & \TA & $0.12{\scriptstyle\pm 0.30}$ & $0.21$ & $0.10{\scriptstyle\pm 0.51}$ & $8.4{\scriptstyle\pm 6.4}$ (waist pitch $\downarrow$) \\
  & \TB & $0.05{\scriptstyle\pm 0.21}$ & $0.02$ & $0.03{\scriptstyle\pm 0.16}$ & $7.8{\scriptstyle\pm 8.7}$ (waist pitch $\downarrow$) \\
  & \TC & $0.05{\scriptstyle\pm 0.22}$ & $0.21$ & $0.03{\scriptstyle\pm 0.16}$ & $7.9{\scriptstyle\pm 8.9}$ (waist pitch $\downarrow$) \\
  & \TD & $0.13{\scriptstyle\pm 0.43}$ & $0.02$ & $0.09{\scriptstyle\pm 0.32}$ & $8.0{\scriptstyle\pm 9.0}$ (waist pitch $\downarrow$) \\
\addlinespace
\multirow{4}{*}{Sit and squat}
  & \TA & $0.53{\scriptstyle\pm 0.51}$ & $1.28$ & $2.18{\scriptstyle\pm 2.22}$ & $19.0{\scriptstyle\pm 7.5}$ (hip $\downarrow$) \\
  & \TB & $0.01{\scriptstyle\pm 0.05}$ & $1.28$ & $0.02{\scriptstyle\pm 0.24}$ & $15.1{\scriptstyle\pm 6.9}$ (hip $\downarrow$) \\
  & \TC & $0.08{\scriptstyle\pm 0.22}$ & $1.28$ & $0.05{\scriptstyle\pm 0.18}$ & $11.4{\scriptstyle\pm 4.7}$ (hip $\downarrow$) \\
  & \TD & $0.81{\scriptstyle\pm 0.80}$ & $1.28$ & $1.88{\scriptstyle\pm 2.19}$ & $13.8{\scriptstyle\pm 5.3}$ (hip $\downarrow$) \\
\addlinespace
\multirow{4}{*}{Kick chair\dag}
  & \TA & $0.01{\scriptstyle\pm 0.08}$ & $0.09$ & $0.01{\scriptstyle\pm 0.12}$ & $0.31{\scriptstyle\pm 0.15}$ (disp. $\downarrow$) \\
  & \TB & $0.01{\scriptstyle\pm 0.08}$ & $0.04$ & $0.01{\scriptstyle\pm 0.18}$ & $0.16{\scriptstyle\pm 0.13}$ (disp. $\downarrow$) \\
  & \TC & $0.00{\scriptstyle\pm 0.05}$ & $0.09$ & $0.00{\scriptstyle\pm 0.06}$ & $0.26{\scriptstyle\pm 0.10}$ (disp. $\downarrow$) \\
  & \TD & $0.01{\scriptstyle\pm 0.06}$ & $0.04$ & $0.01{\scriptstyle\pm 0.09}$ & $0.17{\scriptstyle\pm 0.10}$ (disp. $\downarrow$) \\
\addlinespace
\multirow{4}{*}{Pick up box\dag}
  & \TA & $1.85{\scriptstyle\pm 0.94}$ & 1.89 & $1.73{\scriptstyle\pm 1.18}$ & $0.49{\scriptstyle\pm 0.47}$ (disp. $\downarrow$) \\
  & \TB & $0.36{\scriptstyle\pm 0.68}$ & 0.00 & $0.21{\scriptstyle\pm 0.72}$ & $0.18{\scriptstyle\pm 0.26}$ (disp. $\downarrow$) \\
  & \TC & $0.29{\scriptstyle\pm 0.63}$ & 1.89 & $0.17{\scriptstyle\pm 0.59}$ & $0.20{\scriptstyle\pm 0.24}$ (disp. $\downarrow$) \\
  & \TD & $1.22{\scriptstyle\pm 1.20}$ & 0.00 & $1.17{\scriptstyle\pm 1.40}$ & $0.50{\scriptstyle\pm 0.46}$ (disp. $\downarrow$) \\
\addlinespace
\multicolumn{6}{l}{\dag~Free-object motion; the torque column reports object displacement (disp., m) instead.} \\
\bottomrule
\end{tabular}
\end{table}

\paragraphvA{Per-motion takeaways from \cref{tab:controllability}.}
While the overall trend---contact bodies and impulse track the contact
label---holds across motions, several per-motion patterns are worth
highlighting:
\begin{itemize}[leftmargin=1.5em,itemsep=0pt,topsep=2pt]
  \item \textbf{Sit on table} shows the cleanest separation: when commanded
    contact~\no, both impulse and bodies collapse to near zero (\TB, \TC),
    while contact~\yes recovers them to the reference level (\TA, \TD).
    This is the prototypical case where the contact label fully drives the
    policy's behavior.
  \item \textbf{Step foot on chair} is an interesting partial-control case:
    contact bodies are essentially identical across all four trajectories
    ($\sim\!0.6$), because the robot cannot balance on a single leg with the
    other foot raised that high, so it \emph{must} keep contact with the chair.
    However, the contact impulse still tracks the label (\TC $1.16$\,N$\cdot$s
    vs.\ \TA $1.62$\,N$\cdot$s), showing that the policy modulates how
    \emph{firmly} the foot pushes down even when contact is unavoidable.
  \item \textbf{Lean on backrest I} and \textbf{Wipe whiteboard} show some
    residual contact under contact~\no with near keypoints
    (\TC bodies $0.68$ and $0.52$): the reference keypoints place the
    torso/hand sufficiently close to the surface that fully suppressing contact
    would incur a keypoint tracking error whose cost exceeds the contact-aware
    penalty under our reward weighting, leading the policy to settle at a
    trade-off where residual contact persists.
    The impulse, however, still drops by a factor of $3$--$4\times$.
  \item For the free-object motions, the most informative signal is the
    object displacement. \textbf{Kick chair} only produces a brief
    foot--chair contact (two kicks), so contact bodies and impulse are tiny
    ($\sim\!0.01$) and barely vary across conditions. Yet the displacement
    shows that with contact~\yes and near keypoints (\TA) the two kicks land
    cleanly ($0.31$\,m), while contact~\no (\TC, $0.26$\,m) or far keypoints
    with contact~\yes (\TD, $0.17$\,m) lead to weaker or partial contact.
  \item \textbf{Pick up box} has the largest displacement under contact~\yes
    (\TA $0.49$, \TD $0.50$): the policy commits to grasping and lifting the
    box. Under contact~\no the box is barely displaced (\TB $0.18$, \TC
    $0.20$), even though the keypoint trajectory for \TC is the original
    pickup motion---this is the clearest example of contact conditioning
    enabling task-relevant manipulation that keypoints alone cannot produce.
\end{itemize}

\begin{figure}[!htbp]
\centering
\includegraphics[width=\linewidth]{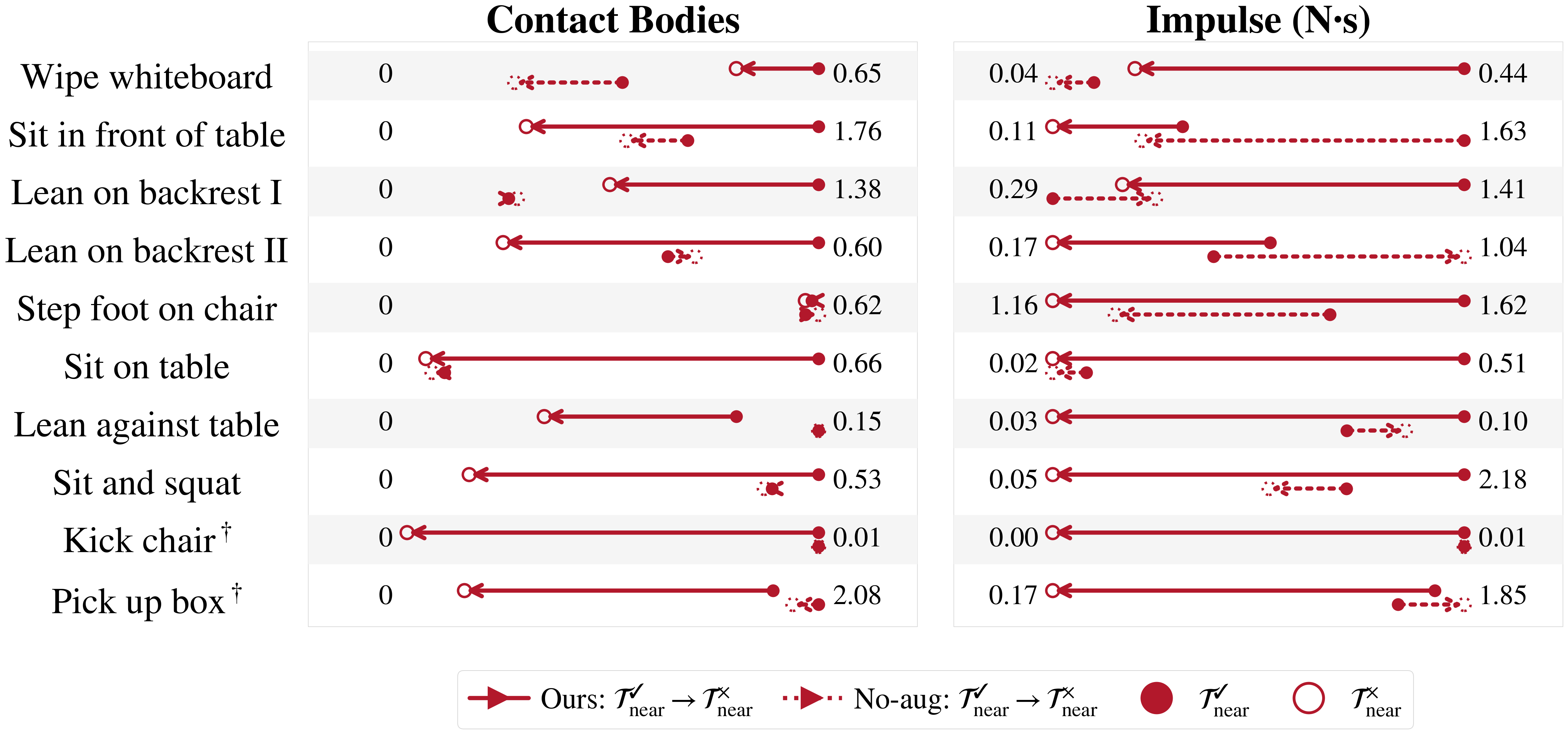}
\captionof{figure}{Per-motion ablation on paired-motion data augmentation, shown for
    both contact bodies and impulse.
    \textcolor[HTML]{b2182b}{Solid arrows} show our full method (Ours);
    \textcolor[HTML]{b2182b}{dotted arrows} show the no-augmentation baseline
    (No-aug). Both arrows go from \TA{} to \TC{}.
    Without augmentation, the arrows are short or even reverse direction,
    indicating the policy fails to follow the contact label.}
\label{fig:ablation_arrows_2col}

\vspace{1em}

\captionof{table}{Ablation on paired-motion data augmentation across all motions.
  (\TA = original keypoint with contact;
  \TC = original keypoint without contact (flipped)).
  Without augmentation, contact metrics are largely insensitive to the contact
  label (both \TA and \TC produce similar values, whether high or low),
  showing the policy ignores the contact conditioning.}
\label{tab:ablation}
\setlength{\tabcolsep}{4pt}
\small
\begin{tabular}{llcccc}
\toprule
\multirow{2}{*}{Motion} & \multirow{2}{*}{Variant}
  & \multicolumn{2}{c}{Contact bodies}
  & \multicolumn{2}{c}{Impulse (N$\cdot$s)} \\
& & \TA & \TC
    & \TA & \TC \\
\midrule
\multirow{2}{*}{Wipe whiteboard}
  & No-aug & $0.34\std{0.40}$ & $0.17\std{0.24}$
           & $0.08\std{0.12}$ & $0.04\std{0.08}$ \\
  & Ours   & $0.65\std{0.45}$ & $0.52\std{0.44}$
           & $0.44\std{0.37}$ & $0.12\std{0.13}$ \\
\addlinespace
\multirow{2}{*}{Sit in front of table}
  & No-aug & $1.20\std{0.44}$ & $0.94\std{0.49}$
           & $1.63\std{1.50}$ & $0.44\std{0.42}$ \\
  & Ours   & $1.76\std{0.33}$ & $0.51\std{0.64}$
           & $0.59\std{0.22}$ & $0.11\std{0.21}$ \\
\addlinespace
\multirow{2}{*}{Lean on backrest I}
  & No-aug & $0.34\std{0.51}$ & $0.37\std{0.40}$
           & $0.29\std{0.66}$ & $0.57\std{1.19}$ \\
  & Ours   & $1.38\std{1.27}$ & $0.68\std{0.65}$
           & $1.41\std{1.46}$ & $0.48\std{0.64}$ \\
\addlinespace
\multirow{2}{*}{Lean on backrest II}
  & No-aug & $0.38\std{0.39}$ & $0.42\std{0.45}$
           & $0.51\std{0.69}$ & $1.04\std{1.42}$ \\
  & Ours   & $0.60\std{0.39}$ & $0.14\std{0.24}$
           & $0.63\std{0.55}$ & $0.17\std{0.37}$ \\
\addlinespace
\multirow{2}{*}{Step foot on chair}
  & No-aug & $0.60\std{0.45}$ & $0.62\std{0.48}$
           & $1.47\std{1.19}$ & $1.23\std{1.02}$ \\
  & Ours   & $0.61\std{0.48}$ & $0.60\std{0.48}$
           & $1.62\std{1.33}$ & $1.16\std{1.01}$ \\
\addlinespace
\multirow{2}{*}{Sit on table}
  & No-aug & $0.06\std{0.17}$ & $0.04\std{0.14}$
           & $0.06\std{0.48}$ & $0.02\std{0.16}$ \\
  & Ours   & $0.66\std{0.47}$ & $0.03\std{0.16}$
           & $0.51\std{0.60}$ & $0.02\std{0.20}$ \\
\addlinespace
\multirow{2}{*}{Lean against table}
  & No-aug & $0.15\std{0.43}$ & $0.15\std{0.43}$
           & $0.08\std{0.23}$ & $0.09\std{0.25}$ \\
  & Ours   & $0.12\std{0.30}$ & $0.05\std{0.22}$
           & $0.10\std{0.51}$ & $0.03\std{0.16}$ \\
\addlinespace
\multirow{2}{*}{Sit and squat}
  & No-aug & $0.47\std{0.50}$ & $0.46\std{0.46}$
           & $1.57\std{1.66}$ & $1.17\std{1.72}$ \\
  & Ours   & $0.53\std{0.51}$ & $0.08\std{0.22}$
           & $2.18\std{2.22}$ & $0.05\std{0.18}$ \\
\addlinespace
\multirow{2}{*}{Kick chair\dag}
  & No-aug & $0.01\std{0.06}$ & $0.01\std{0.07}$
           & $0.01\std{0.27}$ & $0.01\std{0.12}$ \\
  & Ours   & $0.01\std{0.08}$ & $0.00\std{0.05}$
           & $0.01\std{0.12}$ & $0.00\std{0.06}$ \\
\addlinespace
\multirow{2}{*}{Pick up box\dag}
  & No-aug & $2.08\std{0.92}$ & $1.95\std{0.93}$
           & $1.58\std{1.13}$ & $1.85\std{1.26}$ \\
  & Ours   & $1.85\std{0.94}$ & $0.29\std{0.63}$
           & $1.73\std{1.18}$ & $0.17\std{0.59}$ \\
\bottomrule
\end{tabular}
\end{figure}

\paragraphvA{Data augmentation ablation (per-motion).}
\cref{fig:ablation_arrows_2col} visualizes the per-motion arrow plot for both
contact bodies and impulse, and \cref{tab:ablation} reports the underlying
numbers for our method with and without the data augmentation strategies from
\cref{sec:data}, evaluated on \TA and \TC.
Without augmentation, contact metrics are largely insensitive to the contact
label (\TA and \TC produce similar values), showing the policy ignores the
conditioning signal.

\paragraphvA{Per-motion takeaways from \cref{fig:ablation_arrows_2col} and \cref{tab:ablation}.}
The augmentation gap is largest on motions where contact is geometrically
\emph{not forced} by the keypoint reference:
\begin{itemize}[leftmargin=1.5em,itemsep=0pt,topsep=2pt]
  \item \textbf{Sit and squat} shows the most dramatic effect: under contact~\no
    (\TC), No-aug still produces $0.46$ contact bodies and $1.17$\,N$\cdot$s
    impulse (nearly identical to its \TA value of $0.47/1.57$), meaning the
    policy ignores the contact label entirely; with augmentation, the same
    \TC drops to $0.08/0.05$, roughly a $6{-}23\times$ reduction.
  \item \textbf{Sit in front of table}, \textbf{Sit on table}, and \textbf{Lean
    on backrest II} follow the same pattern: No-aug barely modulates contact
    between \TA and \TC, while Ours produces a clear ON/OFF separation.
  \item \textbf{Step foot on chair}, by contrast, has nearly identical numbers
    for No-aug and Ours, because the keypoint reference itself forces foot
    contact regardless of the label---there is little room for augmentation to
    teach contact decoupling here.
  \item \textbf{Pick up box} shows that without augmentation the policy keeps
    grabbing the box even under contact~\no (No-aug \TC: $1.95$ bodies,
    $1.85$\,N$\cdot$s); augmentation cuts these to $0.29/0.17$, allowing the
    same policy to follow a ``don't touch'' command on a manipulation motion.
  \item \textbf{Kick chair} is unaffected by the ablation (both variants $\sim
    0.01$): the contact is so brief and the keypoint trajectory so determined
    by the kick swing that there is essentially no contact-vs-no-contact
    decision for the augmentation to teach.
\end{itemize}

\end{document}